%% file: main.tex
\documentclass[11pt]{article}

\pdfoutput=1
\usepackage{times}
\usepackage{latexsym}
\usepackage[T1]{fontenc}
\usepackage[utf8]{inputenc}

\usepackage{graphicx}
\usepackage{amsmath,amssymb}
\usepackage{iftex}
\usepackage{lmodern}
\usepackage{indentfirst,csquotes}
\usepackage[natbibapa]{apacite}
\usepackage{booktabs}

\usepackage{graphicx}
\usepackage{multirow}
\usepackage{tabularx}
\usepackage{subfig}
\usepackage{lineno}
 \usepackage{amsmath}
\usepackage{booktabs}
\usepackage{paralist}
\usepackage{arydshln}

\usepackage{amssymb,amsthm,amsmath}
\usepackage{xcolor,paralist,hyperref,titlesec,fancyhdr,etoolbox}

\hypersetup{ colorlinks=true, linkcolor=black, filecolor=black, urlcolor=black }

\usepackage{lipsum}
\usepackage{multirow}

\begin{document}
\title{ASR Under the Stethoscope: Evaluating Biases in Clinical Speech Recognition across Indian Languages}\maketitle
\author{\textbf{Subham Kumar\textsuperscript{\dag} Prakrithi Shivaprakash\textsuperscript{\ddag} Abhishek Manoharan\textsuperscript{\ddag} Astut Kurariya\textsuperscript{\ddag} Diptadhi Mukherjee\textsuperscript{$*$} Lekhansh Shukla\textsuperscript{\ddag} Animesh Mukherjee\textsuperscript{\dag} Prabhat Chand\textsuperscript{\ddag} \and Pratima Murthy\textsuperscript{\ddag}} \\
        \textsuperscript{\dag}Indian Institute of Technology, Kharagpur \\ \textsuperscript{\ddag}National Institute of Mental Health and Neuro Sciences, Bangalore \\
        \textsuperscript{$*$} Lokopriya Gopinath Bordoloi Regional Institute of Mental Health, Tezpur \\
        \{kumarshubham209, prakrithishivaprakash, 12.abhishek.m, astutnamo, diptadhimukherjee, drlekhansh, animeshm\}@gmail.com, prabhat@vknnimhans.in, pratimamurthy@gmail.com 
        }

\begin{abstract}
    Automatic Speech Recognition (ASR) is increasingly used to document clinical encounters, yet its reliability in multilingual and demographically diverse Indian healthcare contexts remains largely unknown. In this study, we conduct the first systematic audit of ASR performance on real-world clinical interview data spanning Kannada, Hindi, and Indian English, comparing leading models including IndicWhisper, Whisper-large-v3, Sarvam, Google speech-to-text, Gemma3n, Omnilingual, Vaani, and Gemini. We evaluate transcription accuracy across languages, speakers, and demographic subgroups, with a particular focus on error patterns affecting patients vs. clinicians and gender-based or intersectional disparities. Our results reveal substantial variability across models and languages, with some systems performing competitively on Indian English but failing on code-mixed or vernacular speech. We also uncover systematic performance gaps tied to speaker role and gender, raising concerns about equitable deployment in clinical settings. By providing a comprehensive multilingual benchmark and fairness analysis, our work highlights the need for culturally and demographically inclusive ASR development for India’s healthcare ecosystem.
\end{abstract}

\input{introduction}

\input{related_work}
\input{dataset}
\input{method2}
\input{results}
\input{conclusion}

\bibliographystyle{apacite}
\bibliography{references, manual}


\end{document}

%% file: introduction.tex
\section {Introduction}

The field of psychiatry is highly dependent on language: the clinical interview is the primary diagnostic tool \citet{sommers2015clinical}, unlike other medical disciplines that rely on physical examination, biomarkers and imaging. Therefore, verbatim records of doctor-patient interviews are of immense value for various applications: training, supervision, and teaching in academic setups; data for qualitative, quantitative, linguistic, natural language processing- medical and deep learning research; and reducing the burden of compliance for electronic health record maintenance on clinicians. Despite this utility, generation of manual transcripts for recordings is a significant bottleneck. Manual transcription is highly expensive, labour and time intensive (5-8 hours for 1 hour of audio), prone to fatigue and error, along with inter-annotator variability \citet{mays2019measuring}. Errors in clinical transcripts can alter a clinician’s interpretation of a patient’s condition, a risk noted in early deployments of ASR in psychiatric and medical settings \citep{ciampelli_combining_2023, shikino_clinical_2023}. Automatic speech recognition (ASR) systems are therefore a viable alternative to manual transcription. ASR solutions have expanded significantly, and include proprietary tools like Google Speech-to-text \citet{google_speech_to_text}, Microsoft Azure \citet{azure_speech_to_text}, Amazon Transcribe \citet{amazon_transcribe}, to open source models like OpenAI's Whisper \citet{radford2022whisper}.

While these systems work well for standard English (American and British) and in controlled environments, performance deteriorates substantially for non-standard English, accented English, code-mixing and code-switching which is common in multilingual contexts like India \citet{sitaram2020surveycodeswitchedspeechlanguage}, \citet{koenecke2020racial}. 
Prior work shows that ASR performance degrades substantially for conversational and accented speech \citep{russell_what_2024} and that Indian English remains underrepresented in global training corpora, leading to lower accuracy and greater variability \citep{javed_svarah_2023, rai_deep_2024}. 

The nature of conversation in psychiatric interviews are themselves non-standard. For instance, there is slowness of speech during depression, fast speech during mania, and formal thought disorders like neologisms in psychotic conditions like schizophrenia. Additionally, many domain-specific words and medical jargon may lead to challenges during ASR. Beyond this linguistic complexity, the quality of audio recording itself is often poor: most real-world clinical interviews are conducted in wards and interview rooms, which are not optimised for good quality acoustic recordings, as background noise, noise from machinery and ceiling fans are fairly common. There is also the problem of speaker overlap.

Apart from accuracy, fairness is a critical concern. ASR systems have been shown to exhibit disparities across gender, race, accent, and region \citep{koenecke2020racial, tatman_gender_2017}. Large-scale analyses of Indian English further demonstrate systematic differences in error rates across gender, region, and speech rate \citep{rai_deep_2024}. However, these studies focus on lecture or general conversational speech and do not examine clinical interviews, where speaker roles are unequal socially and interactively. As clinicians and patients assume distinct conversational roles—question-asking versus response-giving —their speech differs in structure, prosody, and the types of expressions. These role-dependent differences could influence ASR recognition patterns. But the extent and direction of such effects in multilingual clinical settings remain understudied.

In this study, we investigate the performance of multiple ASR models in multilingual doctor–patient interviews conducted in Kannada, Hindi, and Indian English. We evaluate the efficiency of different ASR models using WER, CER, and substitution–insertion–deletion analysis, and assess fairness across speaker gender, speaker role (doctor vs. patient), and intersectional categories. By characterising how error patterns differ across languages and speaker groups, we aim to identify where current ASR systems succeed, where they fail, and which forms of clinical communication are most affected. Our goal is to provide an empirical foundation for evaluating ASR in multilingual clinical contexts and to inform the design of systems that are linguistically inclusive, clinically safe, and socially equitable.

Recent advances in multilingual ASR are driven by open-source models like Whisper and IndicWhisper, and by commercial models such as Gemini and Omnilingual, which have integrated speech recognition into multimodal systems. India-focused models such as Sarvam and Vaani address regional languages. Existing evaluations focus on general datasets, lacking comparison in clinical settings, where multilingual, accented, role-dependent speech pose unique challenges. Our study offers the first such comparison. Below is a list of research questions (RQ) that were addressed in this study.

\textbf{RQ1}: How accurately do ASR systems transcribe multilingual clinical interviews across Kannada, Hindi, and Indian English?

\textbf{RQ2}: How does performance differ across ASR models (OpenAI Whisper, IndicWhisper, Sarvam's Saarika-2.5, Google's speech-to-text, Gemma3n, Gemini-2.5-pro, Omnilingual ASR, and Vaani whisper)?

\textbf{RQ3}: How does ASR accuracy vary across languages?

\textbf{RQ4}: Are there systematic performance differences between patient and clinician speech?

\textbf{RQ5}: Are there gender-driven or intersectional disparities in ASR error rates and error patterns?

%% file: related_work.tex
\section{Related Work}
\label{sec:related_work}

Research on ASR in clinical and multilingual settings spans three interconnected areas: A) ASR in clinical communication, B) bias and speaker-level disparities, and C) ASR models’ performance on Indian languages and accents. Each line of work highlights challenges that motivate the present study but leaves important gaps regarding multilingual clinical dialogues and role-based biases.

\subsection{ASR in clinical and psychiatric communication}
A growing set of studies has examined the use of ASR to transcribe or analyze clinical speech. \citet{ciampelli_combining_2023} combined ASR with semantic analysis for schizophrenia-spectrum disorders, reporting substantial error rates and noting that deletions and substitutions disproportionately affected patient speech. \citet{just_moving_2025} demonstrated that ASR distortions alter sentence boundaries and semantic features in clinical interviews, reducing the reliability of subsequent linguistic features. On the other hand, \citet{shikino_clinical_2023} showed that ASR-generated transcripts can support medical trainees’ diagnostic reasoning, even though the accuracy and fairness of the underlying ASR were not evaluated. Another study on Polish medical interviews, documented challenges with clinical terminology, phrasing boundaries, and multiword expressions \cite{kuligowska_challenges_2023}. Collectively, these studies underscore the difficulty of transcribing clinical interactions but remain limited to single languages, do not analyze detailed error types, and do not investigate demographic or interactional sources of bias.

Complementary evidence comes from conversational ASR research. \cite{russell_what_2024} showed that ASR performance on telephone and videoconferencing conversations remains substantially weaker than on read speech, with especially high error rates for short backchannel responses and non-lexical tokens (e.g., “Mmm”, “Hmm”, “Uhh”, “Okay”). Their analysis also revealed systematic variability across speakers, including accents and speaking styles. These findings are directly relevant to clinical interviews, where patients may give brief, disfluent responses and clinicians rely on backchannels to guide the conversation.
\subsection{Bias and speaker-level disparities in ASR}
ASR disparities across demographic groups are increasingly documented. One study identified gender differences in YouTube ASR, driven by phonetic and prosodic variation \citep{tatman_gender_2017}. Another study found that major commercial ASR systems produce nearly twice the error rate for African American speakers compared to white speakers \cite{koenecke2020racial}. In clinical samples, \citet{just_moving_2025} reported higher error rates for speakers born outside Germany, suggesting accent-based inequities even within a single language group. These studies highlight the broader concern that ASR systems are not demographically neutral; however, they do not address multilingual interactions or the socially unequal roles present in clinical interviews.

There remains significant scope for further understanding of how role-based differences influence communication. In healthcare settings, conversation tends to be quite asymmetrical: clinicians typically share longer, well-organized statements, whereas patients often give shorter, gentler, and sometimes uncertain replies. These differences may influence ASR performance in systematic ways, yet no prior study has examined ASR bias across speaker roles or their intersection with gender in real clinical settings.

\subsection{ASR for Indian English and Indian languages}
Recent work has focused on ASR challenges for Indian English and regional Indian languages. Svarah, a benchmark for Indian-accented English, shows that state-of-the-art models such as Whisper exhibit significant accuracy drops and accent-linked variability compared to native English benchmarks \citep{javed_svarah_2023}. \citet{rai_deep_2024} analyzed 8,740 hours of National Programme on Technology Enhanced Learning (NPTEL) lecture speech. They found substantial disparities across gender, region, and speech rate, underscoring the complexity of ASR fairness in Indian contexts. While these studies reveal accent- and demographic-related challenges in Indian English ASR, they focus on monologic or general conversational speech rather than clinical dialogue.

Large-scale multilingual efforts such as IndicSUPERB \citet{javed_indicsuperb_2022} and IndicVoices \citet{javed_indicvoices_2024} provide foundational datasets for Indian speech processing, covering 12-22 Indian languages and thousands of speakers. These resources highlight the linguistic, morphological, and prosodic diversity that complicates ASR in India. They include initial evaluations of gender-related variation in recognition accuracy. However, they do not contain clinical interactions, code-mixed doctor–patient conversations, or analyses of detailed error types. They also do not examine demographic or role-based disparities. On the other hand, the Eka Medical ASR Evaluation Dataset \citep{eka_care_eka_2024} offers valuable Indian accents and drug vocabulary with over 3,900 recordings but is limited to brief, static conversations. It doesn't reflect interactive clinical interviews or assess fairness across gender or role. 

\subsection{Summary and gap}
Across these lines of research, two gaps become clear. First, there is no systematic evaluation of ASR performance in multilingual clinical interviews, despite the interactional complexity
, linguistic diversity, and clinical risks involved. Second, little is known about bias across speaker roles and intersectional categories—e.g., whether ASR systems treat patient speech differently from clinician speech, or whether errors disproportionately affect female patients, male clinicians, or other subgroups. The present study addresses these gaps by auditing ASR accuracy and fairness in Kannada, Hindi, and Indian English doctor–patient interviews, using WER, CER, and substitution–insertion–deletion patterns to characterise how errors differ across languages, genders, roles, and intersectional groups.

%% file: dataset.tex
\section{Dataset}
\label{sec:dataset}
\begin{table}[h!]
    \centering\resizebox{.9\textwidth}{!}{
    \begin{tabular}{cccccc}
        \toprule
        \textbf{Characteristics} & \textbf{Overall} & \textbf{English} & \textbf{Hindi} & \textbf{Kannada} & \textbf{p-value$^\#$} \\
         & N = 162$^*$ & N = 54$^*$ & N = 58$^*$ & N = 50$^*$ \\
        \midrule
        Duration & 32.7 & 36.4 & 26.1  & 36.3  & 0.007 \\
        (minutes) & (12.2, 49.1) & (28.7, 51.1) & (7.6, 44.1) & (4.1, 51.2) &  \\
        Total words & 4896.5 & 5550.5 & 4226.5 & 4361 & 0.007 \\
        & (1747, 7293) & (4121, 7677) & (1205, 7470) & (613, 6869) & \\
        Unique words & 865 & 830 & 730 & 1230 & 0.2 \\
        & (467, 1259) & (599, 1047) & (405, 1162) & (239, 1678) & \\
        Moving average & 0.64 & 0.61 & 0.64 & 0.69 & $<$0.001 \\ 
        type-token ratio & (0.61, 0.67) & (0.59, 0.62) & (0.62, 0.66) & (0.67, 0.71) & \\
        (window=100) &&&&& \\
        \bottomrule
    \end{tabular}}
    \caption{Summary of the dataset. ($^*$) indicates median (Q1, Q3) and ($^\#$) indicates Kruskal-Wallis rank sum test.}
    \label{tab:data_details}
\end{table}
The data for this study comes from a tertiary teaching hospital dedicated to treatment of psychiatric and neurological conditions. There is availability of free inpatient and outpatient treatment for poorer patients and thus a majority of beneficiaries are from such background.
We collected 162 audio recordings of patient and doctor/therapist interactions. These recordings were collected using android mobile phones in mp3 format. While an attempt was made to make recordings in a quiet environment, no special arrangements were made for this. Therefore, the data represents a real-world setting where recordings are made in busy wards and outpatient department rooms. 
The language, duration and lexical diversity of the dataset is summarised in Table ~\ref{tab:data_details} and the speaker profiles are detailed in Table ~\ref{tab:speaker_profile}.
\begin{table}[!t]
    \centering\resizebox{.75\textwidth}{!}{
    \begin{tabular}{cccccc}
    \toprule
        \textbf{Characteristics} & \textbf{Overall} & \textbf{English} & \textbf{Hindi} & \textbf{Kannada} & \textbf{$p$-value$^\#$} \\
         & N = 162$^*$ & N = 54$^*$ & N = 58$^*$ & N = 50$^*$ \\
        \midrule
        \textbf{Patient's Age} & 31.5 & 36 & 30 & 30.5 & 0.7 \\
        & (26, 42) & (26, 39) & (26, 45) & (26, 42) & \\
        \textbf{Doctor's Age} & 29 & 27 & 29 & 30 & $<$0.001 \\
        & (27, 30) & (27, 27) & (28, 31) & (30, 31) & \\
        \midrule
        \textbf{Patient's Gender} &&&&&\\
        F & 31 & 2 & 10 & 19 & $<$0.001 \\
        & (19.1\%) & (3.7\%) & (17.2\%) & (38\%) & \\
        M & 131 & 52 & 48 & 31 & $<$0.001 \\
        & (80.8\%) & (96.3\%) & (82.7\%) & (62\%) & \\
        \midrule
        \textbf{Doctor's Gender} &&&&&\\
        F & 104 & 54 & 30 & 20 & $<$0.001 \\
        & (64.2\%) & (100\%) & (51.7\%) & (40\%) & \\
        M & 58 & 0 & 28 & 30 & $<$0.001 \\
        &  (35.8\%) & (0\%) & (48.3\%) & (60\%) & \\
        \midrule
        \textbf{Patient's Education Level} &&&&& \\
        \textbf{Graduate} & 50 & 34 & 7 & 9 & $<$0.001 \\
        & (30.9\%) & (62.9\%) & (12.1\%) & (18\%) & \\
        \textbf{Not formally educated} & 3 & 0 & 3 & 0 & $<$0.001 \\
        & (1.8\%) & (0\%) & (5.2\%) & (0\%) & \\
        \textbf{Post Graduate} & 10 & 2 & 7 & 1 & $<$0.001 \\
        & (6.2\%) & (3.7\%) & (12.1\%) & (2\%) & \\
        \textbf{Primary} & 31 & 0 & 13 & 18 & $<$0.001 \\
        & (19.1\%) & (0\%) & (22.4\%) & (36\%) & \\
        \textbf{Secondary} & 68 & 18 & 28 & 22 & $<$0.001 \\
        & (42\%) & (33.3\%) & (48.3\%) & (44\%) & \\
        \midrule
        \textbf{Patient's Country Region} &&&&&\\
        \textbf{Central} & 2 & 0 & 2 & 0 & 0.005 \\
        & (1.2\%) & (0\%) & (3.4\%) & (0\%) & \\ 
        \textbf{East} & 7 & 3 & 4 & 0 & 0.005 \\
        & (4.3\%) & (5.5\%) & (6.9\%) & (0.00\%) & \\
        \textbf{North} & 9 & 3 & 6 & 0 & 0.005 \\
        & (5.5\%) & (5.5\%) & (10.3\%) & (0\%) & \\
        \textbf{Northeast} & 8 & 5 & 3 & 0 & 0.005 \\
         & (4.9\%) & (9.2\%) & (5.2\%) & (0\%) & \\
        \textbf{South} & 134 & 42 & 42 & 50 & 0.005 \\
         & (82.7\%) & (77.8\%) & (72.4\%) & (100\%) & \\
        \textbf{West} & 2 & 1 & 1 & 0 & 0.005 \\
         & (1.2\%) & (1.8\%) & (1.7\%) & (0\%) & \\
         \midrule
        \textbf{Doctor's Country Region}\\
        \textbf{Central} & 28 & 0 & 28 & 0 & $<$0.001\\
        & (17.3\%) & (0\%) & (48.3\%) & (0\%) & \\
        \textbf{East} & 25 & 12 & 12 & 1 & $<$0.001 \\
        & (15.4\%) & (22.2\%) & (20.7\%) & (2\%) & \\
        \textbf{South} & 109 & 42 & 18 & 49 & $<$0.001 \\
        & (67.3\%) & (77.8\%) & (31\%) & (98\%) & \\
        \bottomrule
    \end{tabular}}
    \caption{Summary of speaker profiles. ($*$) Median (Q1, Q3), ($^\#$) Kruskal-Wallis rank sum test.}
    \label{tab:speaker_profile}
\end{table}

This dataset contains speech from 117 unique speakers, including 7 doctors/therapists and 110 patients. All conversations are between two individuals, making 112 unique doctor-patient/therapist-patient pairs.

\subsection{Preprocessing}
All recordings were listened to by two psychiatrists to ensure they were intelligible. It was ensured that the recordings did not contain the name of the patient or any numerical identifier like phone number, etc. However, we did not exclude segments that contained names of places, dates, etc, as we wish to evaluate if such named entities lead to more errors in ASR.

\subsection{Annotation}
As part of earlier research, transcripts in native languages were available for 103 of these recordings. For the remaining transcripts (60), two psychiatrists transcribed the recordings. The annotation guidelines are given in the Appendix.

%% file: method2.tex
\section{Method}
\label{sec:method}
In this section, we describe the methodology of the study and the evaluation methods that were employed to assess the findings. The procedure for compiling the ASR-generated transcripts is first discussed, followed by a definition of the word error rate (WER) metric and the statistical tests that were performed to ascertain the differences between various ASR models. We evaluate a total of 8 ASR models, including one large language model \textsc{Gemini-2.5-pro}. 
\subsection{Methodology}
Recall that we have the audio files in an interview format where there are exactly two speakers, i.e., the patient and the doctor. We generate transcripts for each of these audio files with word level timestamps and a standalone transcript as a whole. Four of these  ASR models (OpenAI Whisper, IndicWhisper, Sarvam's Saarika-2.5, and Google's speech-to-text) output word-level timestamps with transcription text, while the remaining four models (Gemma3n, Gemini-2.5-pro, Omnilingual ASR, and Vaani whisper) generate only the transcription text.

\subsection{ASR models}
We briefly describe the ASR model used to transcribe long audio files.
\begin{itemize}
    \item \textbf{Whisper.} Whisper large-v3 is a 1.55B parameter transformer encoder–decoder ASR model that uses windowed inference to scale to arbitrarily long audio segments based on 30-second log-Mel spectrogram segments \citep{openai_whisper_large_v3}. 
    \item \textbf{IndicWhisper.} It uses a transformer-based encoder–decoder architecture that was modified from Whisper and enhanced with multilingual acoustic–textual alignment and subword tokenization tailored for Indic scripts \citep{indicwhisper2024}. 
    \item \textbf{Sarvam.} With long-form processing using windowed inference, Sarvam's Saarika-2.5 is a large-scale multilingual ASR model tailored for Indian languages that shares architectural similarities with Whisper-style encoder–decoder transformers. It builds robust acoustic representations by feeding log-Mel (or similar spectral) characteristics into a deep self-attention encoder \citep{sarvam2025}. 
    \item \textbf{Gemma3n.} Google's Gemma3n is an effective multimodal generative AI model that uses the Matryoshka transformer (MatFormer) architecture with nested sub-models for scalable computing, which can be deployed on devices with limited resources \citep{gemmateam2025gemma3technicalreport}.  For long-form transcription, it combines segments using timestamps and confidence metrics after processing 30-second audio segments with streaming support for speech recognition. 
    \item \textbf{Google speech-to-text.} The Chirp-3 generative model powers Google's speech-to-text ASR, which uses a multilingual neural architecture with deep transformer-based encoders to transform log-Mel spectrograms from 16kHz audio into rich acoustic embeddings \citep{google_speech_to_text}. 
    \item \textbf{Gemini-2.5-pro.} Gemini-2.5-Pro integrates a multimodal transformer–based ASR pipeline that converts log-Mel spectral features into semantic representations employing cross-modal attention. Its architecture is designed for long contexts. The architecture enables reliable transcription of extended conversational audio \citep{gemini25pro_2025}. 
    \item \textbf{Omnilingual.} Meta’s Omnilingual ASR is a multilingual speech recognition system that supports over 1,600 languages through a unified transformer-based encoder–decoder architecture. It uses self-supervised pre-training on a large speech encoder with up to 7 billion parameters. Afterwards, the model is fine-tuned with a decoder-based ASR system, trained on 4.3M+ hours of diverse audio. This approach guarantees reliable speech recognition even for low-resource languages and makes it easier to add new languages with only a small amount of data \citep{omnilingualasrteam2025omnilingualasropensourcemultilingual}. We used the \texttt{omniASR-LLM-7B} variant for our experiments. 
    \item \textbf{Vaani Whisper.} IISc ARTPARK's Vaani Whisper ASR models are refined versions of OpenAI's Whisper-medium/large (e.g., \texttt{whisper-medium-vaani-hindi}, and \texttt{whisper-medium-vaani-kannada}) on more than 700 hours of Indian voice data for reliable multilingual transcription. For our experiments, we have used \texttt{ARTPARK-IISc/whisper-large-v3-vaani-hindi} \citet{vaani_whisper_hi_2025} and \texttt{ARTPARK-IISc/whisper-medium-vaani-kannada} \cite{vaani_whisper_kn_2025}, both being the largest for these two languages. 
\end{itemize}

\subsection{Evaluation metric}
The main indicator used to assess the accuracy of ASR transcription is the word error rate (WER). Since WER is the most widely used metric for assessing ASRs and has been utilized by several researchers in the literature, we have chosen it as the evaluation metric. WER is calculated mathematically by dividing the total number of words (N) in the reference transcript by the sum of substitutions (S), deletions (D), and insertions (I).
\[
\text{WER \%} = \frac{S + D + I}{N}*100
\]
It measures the percentage of words incorrectly recognized by the ASR system relative to a ground truth transcript. First, we normalize both the ground truth and the generated transcripts by preproceesing the text. This preprocessing includes lowercasing, removing punctuation, and standardizing numbers to reduce superficial mismatches. WER measures ASR performance in terms of word recognition fidelity by taking word-level errors into account. Lower WER indicates better accuracy. Second, we use the JiWER Python library for the calculation of WER.

%% file: results.tex
\section{Results}
\label{sec:results}
\begin{table}[!ht]
    \centering\centering\resizebox{.75\textwidth}{!}{
    \begin{tabular}{cccc}
    \toprule
         \textbf{Models} & \textbf{languages} & \textbf{WER( in \%)} & \textbf{(S, D, I) \%}  \\
         \midrule
         \multirow{3}{*}{Whisper} & English & 46.76 & (9.48, 19.81, 21.76) \\
         & Hindi & 71.68 & (26.23, 39.17, 9.21) \\
         & Kannada & 98.55 & (22.22, 76.51, 0.0014) \\
         \midrule
         \multirow{3}{*}{IndicWhisper} & English & - & - \\
         & Hindi & 70.3 & (23.44, 45.58, 4.27) \\
         & Kannada & 97.05 & (29.03, 68.21, 0.02) \\
         \midrule
         \multirow{3}{*}{Sarvam} & English & 34.33 & (8.94 ,16.40, 36.87) \\
         & Hindi & 39.03 & (14.47, 6.76, 39.15) \\
         & Kannada & 54.37 & (27.87, 17.68, 39.03) \\
         \midrule
         \multirow{3}{*}{GSTT}  & English & 74.60 & (36.13, 40.22, 9.71) \\
         & Hindi & 85.55 & (10.09, 75.60, 0.13) \\
         & Kannada & 94.90 & (11.34, 84.01, 0.005) \\
         \midrule
         \multirow{3}{*}{Gemma3n} & English & 40.22 & (14.63, 17.64, 19.54) \\
         & Hindi & 48.14 & (19.88, 6.45, 26.48) \\
         & Kannada & 90.90 & (48.57, 16.67, 30.47) \\
         \midrule
         \multirow{3}{*}{Gemini} & English & 14.15 & (5.28, 18.58, 4.94) \\
         & Hindi & 18.52 & (11.32, 4.35, 8.74) \\
         & Kannada & 35.01 & (22.71, 16.86, 7.53) \\
         \midrule
         \multirow{3}{*}{Omnilingual} & English & 58.64 & (23.40, 31.20, 4.00) \\
         & Hindi & 44.42 & (21.54, 15.60, 6.29) \\
         & Kannada & 77.21 & (5.56, 24.92, 1.73) \\
         \midrule
         \multirow{3}{*}{Vaani Whisper} & English & - & - \\
         & Hindi & 43.55 & (23.76, 12.20, 6.53) \\
         & Kannada & 75.35 & (41.08, 32.19, 2.06) \\
    \bottomrule
    \end{tabular}}
    \caption{Overall performance of ASR models across all 3 languages (English, Hindi, and Kannada) under study. S, D, I(\%) column indicates median of insertion, deletion and substitution component of WER corresponding to each subgroup.}
    \label{tab:overall_perf}
\end{table}

In this section, we present the results of our experimental evaluation of 8 ASR models for the transcription task across 3 languages (English, Hindi, and Kannada). First, we evaluated the overall performance for each of the 8 models in all 3 languages supported by their substitution, deletion, and insertion error rates. Then we report the performance of the best ASR model with word-level timestamps on each of the characteristic attributes mentioned in Table ~\ref{tab:speaker_profile}.

\subsection{Overall performance}
The Table~\ref{tab:overall_perf} reports the WER and its constituent parts, substitution (S), deletion (D), and insertion (I) in percentage, to compare ASR models in three languages, namely English, Hindi, and Kannada. As we observe from this table, Gemini got relatively lower WER scores (English: 14.15\%, Hindi: 18.52\%, Kannada: 35.01\%), demonstrating comparatively better multilingualism and generalization. 
Among the three languages, Kannada poses relatively higher challenge to the ASR models because of the lack of clear word boundaries in speech,  complex conjugates and regional accents. Further, Hindi being a moderate resource and Kannada being a comparatively low resource justifies the high WER scores among models. Lastly, models like Whisper, GSTT, and Gemma3n show a very high WER for Kannada because their architectures and training corpora are not deeply optimised for rich Indian phonetic diversity and retroflex phonemes in Dravidian languages. 

We can observe that for the Kannada language, the error mostly stems from deletions (D) reaching as high as 84\% for GSTT. The deletions possibly result from the ASR models failing to recognise complex conjugates. Sarvam has an exceptionally high insertion rate (I) in both Hindi (39.15\%) and English (36.87\%). Usually, this indicates hallucinations. Neural ASR models may be caught in a loop (repeating ``and... and... and...'') or produce fluent but inaccurate text based on their internal language model rather than the audio when they are unsure or come across quiet or noise. This increases the number of ``ghost words'', which increases the insertion rate.
\if{0}\begin{figure}
    \centering
    \includegraphics[width=0.5\linewidth]{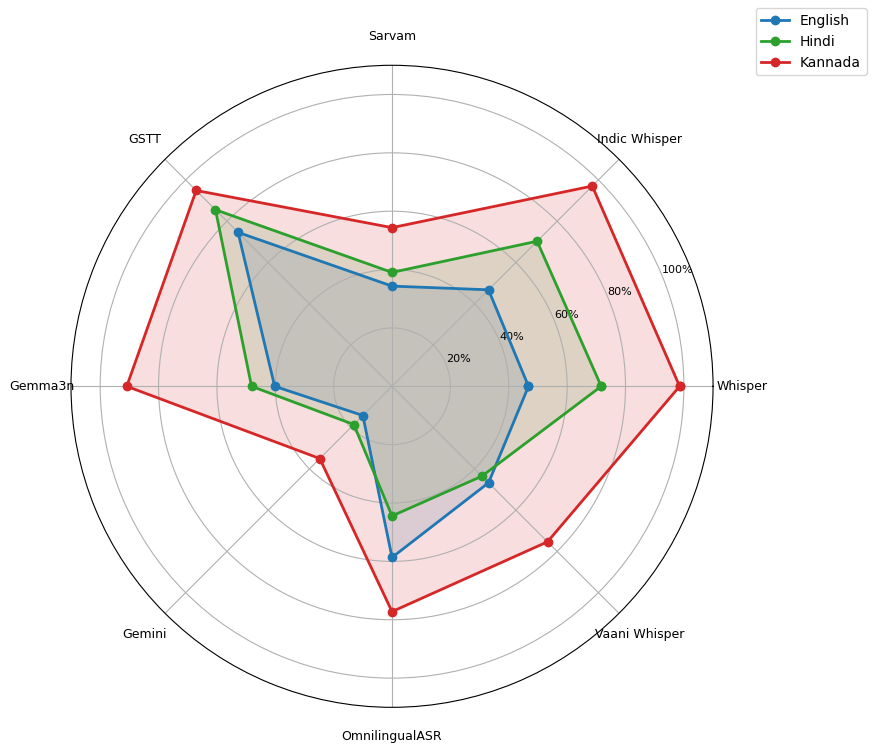}
    \caption{WER plot of ASR models across languages.}
    \label{fig:wer_radarplot}
\end{figure}\fi

\if{0}\begin{figure}
    \centering
    \includegraphics[width=0.5\linewidth]{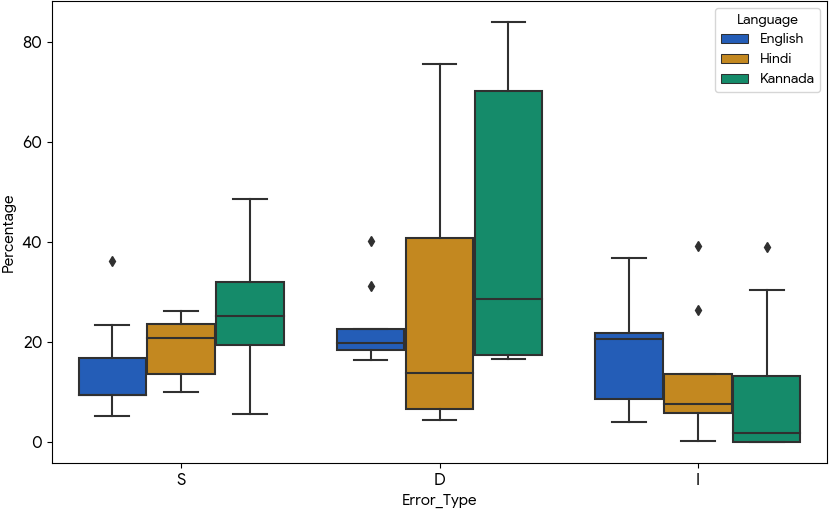}
    \caption{Comparative analysis of error types across languages in ASR models.}
    \label{fig:sid_boxplot}
\end{figure}\fi

\subsection{Performance of ASR models on attributes}
We will now compare the performance of the best ASR model (\textsc{Sarvam}), which gave the word level timestamps across different attributes for each of the three languages as reported in Table~\ref{tab:wer_role}.

\noindent\textbf{Gender. } We compare the male and female WER for each of the languages. We observe from Table~\ref{tab:wer_role} that WER is slightly greater for male than female considering all the languages combined together. This generally happens because the frequency of male voice is close to noise i.e., 300Hz. For females it is generally higher as mentioned in ~\citep{denoasr}. However, for Hindi, we observe that the female WER is relatively higher than the male WER and this behaviour ($\Delta <$ 0, where $\Delta$ = maleWER - femaleWER) is observed across all whisper based ASR models with $\Delta$ ranging from 2-10\%. 

\begin{figure}
    \centering
    \includegraphics[width=0.5\linewidth]{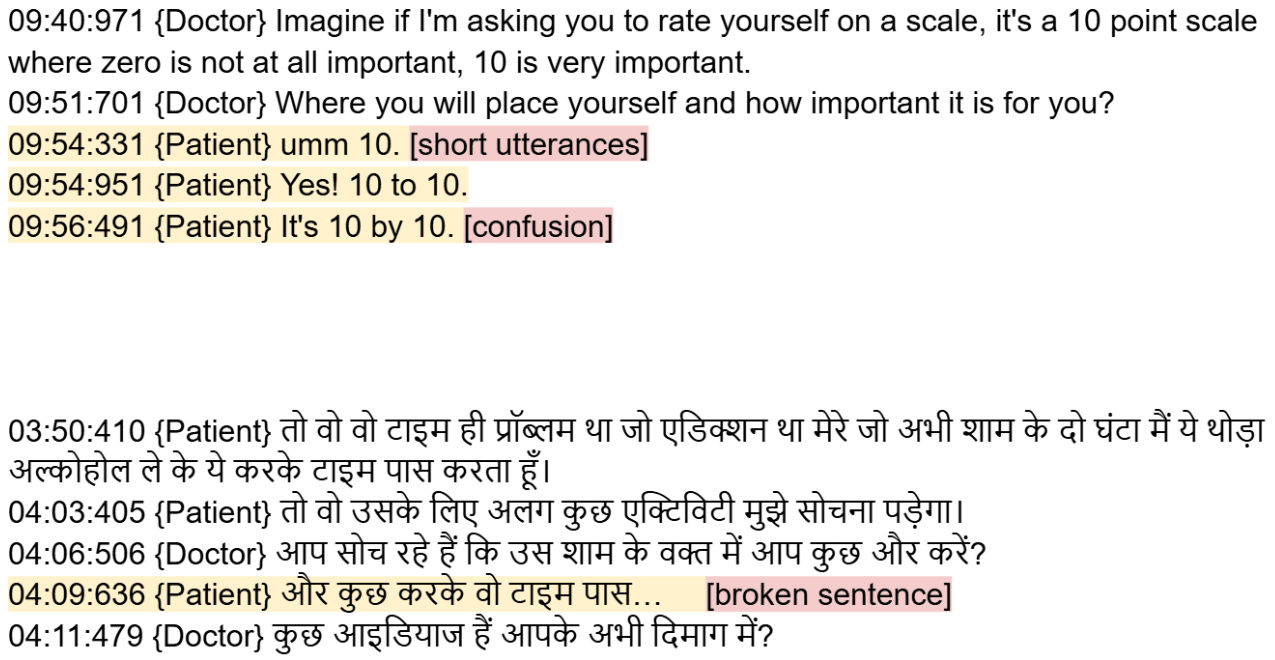}
    \caption{Real examples of patient-doctor conversion showing caveats in patient dialogue.}
    \label{fig:diag}
\end{figure}

\noindent\textbf{Speaker role. } 
A Mann-Whitney U test revealed a statistically significant difference in WER between speaker roles (p $<$ 0.001). Patients showed a significantly higher median WER (56.75) compared to doctors (50.45) indicates a small-to-moderate negative association, suggesting that doctors are associated with lower error rates. Chiu et al. provide separate quantitative WER for both doctors and patients \citep{chiu2017_medical_asr}.
Some other studies have also found higher WER among patients than among health care providers, but they did not specify exact percentages \citep{miner2020_asr_psychotherapy, tran2022_digital_scribes}. They say that patient speech is hard for ASR systems because it is often spontaneous and disfluent, has more varied accents, and includes short utterances or broken-up sentences as shown in Figure ~\ref{fig:diag}. Several other factors may have driven the high WER difference in our dataset. First, there are acoustic issues: microphone placement, background noise, and high overlap rates often affect patient segments. Second, the linguistic choices differ. While doctors stick to predictable medical phrasing, patients may use local vocabulary and code-switching. To explore it further, we need acoustic adaptation and error analysis.

\noindent\textbf{Gender $\times$ Speaker. } Here we report WER at the intersection of gender and role of the speaker. Significant disparities are observed in these intersectional groups. The female doctor has significantly lower WER than other attributes followed by female patient, then male doctor and highest for male patient. This follows the general trajectory of both Gender and Speaker role individually. A Mann-Whitney U test revealed a statistically significant difference in WER within gender $\times$ speaker roles (p $<$ 0.01).

\noindent\textbf{Education level. } The education level of patient is broadly classified into two categories including ``below graduation'' and ``graduation or above''. This is done to tackle the problem of skewness when the education level is categorized into fine-grained classes as presented in Table ~\ref{tab:data_details}. The patient group with education level  ``graduation or above'' exhibits significantly lower WER (37.57\%) than the patient group with ``below graduation'' education level (43.15\%). While we found that lower speakers' education linked to higher error rates, direct evidence from prior ASR research is limited. Most existing studies examine demographic factors such as accent, dialect, native/non-native speech, or racial/ethnic background as predictors of ASR performance \citep{nakatumba2025_asr_bias_review, cunningham2025_asr_bias_equity}. We think the problem has two main parts. First, people with less education often speak in ways (for example using strong dialects or colloquialisms) that aren’t well represented in the standard datasets. Second, and more importantly, speech recognition models are often trained on perfect, scripted, or read-aloud speech. These models struggle when dealing with lingua-franca speech. Therefore, high error rates might not mean that the speakers are hard to understand, but that the models are too delicate and only work well with their ideal training data.

\noindent\textbf{Native Region. } WER is impacted by regional variance because of prosodic rhythm variations, phoneme alterations and accents. Certain regions, like the Northeast and North, have the lowest WER because their accents may have clearer enunciation patterns. Other regions like West and English-speaking Central regions see high WER due to stronger regional accents, vowel elongations, dropped consonants, or prosodic emphasis patterns unfamiliar to the model. Strangely, the WER for English-speaking patients from West region is as high as 60.76\%. This indicates that patients from West region use abbreviations or short forms very often, which is likely to inflate the WER. For Hindi language, the Northeast region achieves the best WER followed by East, West, and Southern regions. The count of files for patient region is tabulated in Table ~\ref{tab:speaker_profile}.

\begin{table}
    \centering\centering\resizebox{.75\textwidth}{!}{
    \begin{tabular}{cccccc}
    \toprule
    \textbf{Category} & \textbf{Attribute} & S-Hindi & S-English & S-Kannada & S-Overall \\
    \midrule
    \multirow{2}{*}{\textbf{Gender}} & \textbf{Male} & 46.50 & 47.27 & 59.78 & 50.98 \\
    & \textbf{Female} & 56.20 & 44.44 & 49.92 & 47.31 \\
    \midrule
    \multirow{2}{*}{\textbf{Speaker role}} & \textbf{Patient} & 56.20 & 47.22 & 72.13 & 56.75 \\
    & \textbf{Doctor} & 45.38 & 44.44 & 62.56 & 50.45 \\
    \midrule
    \multirow{4}{*}{\textbf{Gender \& Role}} & \textbf{Male Patient} & 62.43 & 59.62 & 71.92 & 63.19 \\
    & \textbf{Male Doctor} & 56.67 & - & 76.30 & 65.60 \\
    & \textbf{Female Patient} & 54.96 & 47.94 & 71.92 & 62.60 \\
    & \textbf{Female Doctor} & 45.18 & 41.40 & 66.63 & 48.42 \\
    \midrule
    \multirow{2}{*}{\textbf{Education}} & \textbf{Below Graduation} & 39.40 & 31.90 & 54.27 & 43.15 \\
    & \textbf{Graduation or above} & 37.06 & 32.77 & 48.86 & 37.57 \\
    \midrule
    \multirow{6}{*}{\textbf{Native Region}} & \textbf{Central} & 46.65 & - & - & 46.65 \\
    & \textbf{East} & 39.52 & 30.79 & - & 31.24 \\
    & \textbf{North} & 43.47 & 37.99 & - & 37.99 \\
    & \textbf{Northeast} & 31.90 & 38.18 & - & 36.01 \\
    & \textbf{South} & 39.36 & 32.59 & 54.37 & 44.16 \\
    & \textbf{West} & 39.52 & 60.76 & - & 57.87 \\
    \bottomrule
    \end{tabular}}
    \caption{Attribute wise WER\% of \textsc{Sarvam} model across three languages. S-Hindi, S-English, S-Kannada, and S-Overall indicates Sarvam's performance on Hindi, English, Kannada, and all the three languages combined, respectively.}
    \label{tab:wer_role}
\end{table}

%% file: conclusion.tex
\section{Conclusion}
\label{sec:conclusion}

In this study, we highlighted the disparities in ASR models, particularly in clinical psychiatric interview setting between patient and doctor. We comprehensively audit eight state-of-the-art models across three linguistically diverse languages with Indian speakers from diverse geographical regions of India in terms of WER. In addition, we compare the substitution, deletion and insertion error rates in WER to analyze the most frequent component which resulted in such WER scores. Finally, we investigate the WER performance  of \textsc{Sarvam} across different categories.

Our findings suggest that almost all models have WER scores lowest for English, highest for Kannada, with Hindi in the middle (except for Omnilingual ASR). However, this may occur due to the biasness in their training data. The WER scores for each language across models is considered moderate to high which suggests the need to develop an ASR model for clinical psychiatric interview setting, trained from scratch.

%% file: manual.bib
@misc{google_speech_to_text,
  author = {Google Cloud},
  year   = {2023},
  title  = {Speech-to-Text Documentation},
  howpublished = {\url{https://cloud.google.com/speech-to-text}},
  note   = {Accessed 2025-11-21}
}

@misc{azure_speech_to_text,
  author       = {Microsoft Azure},
  year         = {2023},
  title        = {Azure Speech-to-Text Documentation},
  howpublished = {\url{https://learn.microsoft.com/azure/ai-services/speech-service/}},
  note         = {Accessed 2025-11-21}
}

@misc{amazon_transcribe,
  author       = {Amazon Web Services},
  year         = {2023},
  title        = {Amazon Transcribe Documentation},
  howpublished = {\url{https://docs.aws.amazon.com/transcribe/}},
  note         = {Accessed 2025-11-21}
}

@article{radford2022whisper,
  title        = {Robust Speech Recognition via Large-Scale Weak Supervision},
  author       = {Radford, Alec and Kim, Jong Wook and Xu, Tao and Brockman, Greg and McLeavey, Christine and Sutskever, Ilya},
  journal      = {arXiv preprint arXiv:2212.04356},
  year         = {2022},
  url          = {https://arxiv.org/abs/2212.04356},
  note         = {Accessed 2025-11-29}
}

@misc{sitaram2020surveycodeswitchedspeechlanguage,
      title={A Survey of Code-switched Speech and Language Processing}, 
      author={Sunayana Sitaram and Khyathi Raghavi Chandu and Sai Krishna Rallabandi and Alan W Black},
      year={2020},
      eprint={1904.00784},
      archivePrefix={arXiv},
      primaryClass={cs.CL},
      url={https://arxiv.org/abs/1904.00784}, 
}

@article{sommers2015clinical,
  title={The clinical interview},
  author={Sommers-Flanagan, John and Zeleke, Waganesh Abeje and Hood, ME},
  journal={The encyclopedia of clinical psychology. John Wiley \& Sons, Inc, Hoboken},
  pages={1--9},
  year={2015}
}

@article{mays2019measuring,
  title={Measuring the rate of manual transcription error in outpatient point-of-care testing},
  author={Mays, James A and Mathias, Patrick C},
  journal={Journal of the American Medical Informatics Association},
  volume={26},
  number={3},
  pages={269--272},
  year={2019},
  publisher={Oxford Academic}
}

@article{koenecke2020racial,
  title={Racial disparities in automated speech recognition},
  author={Koenecke, Allison and Nam, Andrew and Lake, Emily and Nudell, Joe and Quartey, Minnie and Mengesha, Zion and Toups, Connor and Rickford, John R and Jurafsky, Dan and Goel, Sharad},
  journal={Proceedings of the national academy of sciences},
  volume={117},
  number={14},
  pages={7684--7689},
  year={2020},
  publisher={National Academy of Sciences}
}

@misc{openai_whisper_large_v3,
  author       = {OpenAI},
  year         = {2024},
  title        = {Whisper Large-v3 Model Card},
  howpublished = {\url{https://platform.openai.com/docs/models/whisper}},
  note         = {A 1.55B-parameter transformer encoder--decoder ASR model using windowed inference on 30-second log-Mel spectrogram segments. Accessed 2025-11-21}
}

@misc{indicwhisper2024,
  author       = {AI4Bharat},
  year         = {2024},
  title        = {IndicWhisper: Multilingual ASR model for Indian languages},
  howpublished = {\url{https://ai4bharat.iitm.ac.in/areas/asr}},
  note         = {A transformer-based encoder--decoder architecture modified from Whisper and enhanced with multilingual acoustic--textual alignment and subword tokenization tailored for Indic scripts. Accessed 2025-11-21}
}

@misc{sarvam2025,
  author       = {SarvamAI},
  year         = {2025},
  title        = {Sarvam AI Speech-to-Text API Documentation},
  howpublished = {\url{https://docs.sarvam.ai/api-reference-docs/api-guides-tutorials/speech-to-text/overview}},
  note         = {Accessed 2025-11-21. Saarika / Saaras ASR models support long-form, multilingual speech recognition with windowed inference and log-Mel input features.}
}

@misc{gemmateam2025gemma3technicalreport,
      title={Gemma 3 Technical Report}, 
      author={Gemma Team and Aishwarya Kamath and Johan Ferret and Shreya Pathak and Nino Vieillard and Ramona Merhej and Sarah Perrin and Tatiana Matejovicova and Alexandre Ramé and Morgane Rivière and Louis Rouillard and Thomas Mesnard and Geoffrey Cideron and Jean-bastien Grill and Sabela Ramos and Edouard Yvinec and Michelle Casbon and Etienne Pot and Ivo Penchev and Gaël Liu and Francesco Visin and Kathleen Kenealy and Lucas Beyer and Xiaohai Zhai and Anton Tsitsulin and Robert Busa-Fekete and Alex Feng and Noveen Sachdeva and Benjamin Coleman and Yi Gao and Basil Mustafa and Iain Barr and Emilio Parisotto and David Tian and Matan Eyal and Colin Cherry and Jan-Thorsten Peter and Danila Sinopalnikov and Surya Bhupatiraju and Rishabh Agarwal and Mehran Kazemi and Dan Malkin and Ravin Kumar and David Vilar and Idan Brusilovsky and Jiaming Luo and Andreas Steiner and Abe Friesen and Abhanshu Sharma and Abheesht Sharma and Adi Mayrav Gilady and Adrian Goedeckemeyer and Alaa Saade and Alex Feng and Alexander Kolesnikov and Alexei Bendebury and Alvin Abdagic and Amit Vadi and András György and André Susano Pinto and Anil Das and Ankur Bapna and Antoine Miech and Antoine Yang and Antonia Paterson and Ashish Shenoy and Ayan Chakrabarti and Bilal Piot and Bo Wu and Bobak Shahriari and Bryce Petrini and Charlie Chen and Charline Le Lan and Christopher A. Choquette-Choo and CJ Carey and Cormac Brick and Daniel Deutsch and Danielle Eisenbud and Dee Cattle and Derek Cheng and Dimitris Paparas and Divyashree Shivakumar Sreepathihalli and Doug Reid and Dustin Tran and Dustin Zelle and Eric Noland and Erwin Huizenga and Eugene Kharitonov and Frederick Liu and Gagik Amirkhanyan and Glenn Cameron and Hadi Hashemi and Hanna Klimczak-Plucińska and Harman Singh and Harsh Mehta and Harshal Tushar Lehri and Hussein Hazimeh and Ian Ballantyne and Idan Szpektor and Ivan Nardini and Jean Pouget-Abadie and Jetha Chan and Joe Stanton and John Wieting and Jonathan Lai and Jordi Orbay and Joseph Fernandez and Josh Newlan and Ju-yeong Ji and Jyotinder Singh and Kat Black and Kathy Yu and Kevin Hui and Kiran Vodrahalli and Klaus Greff and Linhai Qiu and Marcella Valentine and Marina Coelho and Marvin Ritter and Matt Hoffman and Matthew Watson and Mayank Chaturvedi and Michael Moynihan and Min Ma and Nabila Babar and Natasha Noy and Nathan Byrd and Nick Roy and Nikola Momchev and Nilay Chauhan and Noveen Sachdeva and Oskar Bunyan and Pankil Botarda and Paul Caron and Paul Kishan Rubenstein and Phil Culliton and Philipp Schmid and Pier Giuseppe Sessa and Pingmei Xu and Piotr Stanczyk and Pouya Tafti and Rakesh Shivanna and Renjie Wu and Renke Pan and Reza Rokni and Rob Willoughby and Rohith Vallu and Ryan Mullins and Sammy Jerome and Sara Smoot and Sertan Girgin and Shariq Iqbal and Shashir Reddy and Shruti Sheth and Siim Põder and Sijal Bhatnagar and Sindhu Raghuram Panyam and Sivan Eiger and Susan Zhang and Tianqi Liu and Trevor Yacovone and Tyler Liechty and Uday Kalra and Utku Evci and Vedant Misra and Vincent Roseberry and Vlad Feinberg and Vlad Kolesnikov and Woohyun Han and Woosuk Kwon and Xi Chen and Yinlam Chow and Yuvein Zhu and Zichuan Wei and Zoltan Egyed and Victor Cotruta and Minh Giang and Phoebe Kirk and Anand Rao and Kat Black and Nabila Babar and Jessica Lo and Erica Moreira and Luiz Gustavo Martins and Omar Sanseviero and Lucas Gonzalez and Zach Gleicher and Tris Warkentin and Vahab Mirrokni and Evan Senter and Eli Collins and Joelle Barral and Zoubin Ghahramani and Raia Hadsell and Yossi Matias and D. Sculley and Slav Petrov and Noah Fiedel and Noam Shazeer and Oriol Vinyals and Jeff Dean and Demis Hassabis and Koray Kavukcuoglu and Clement Farabet and Elena Buchatskaya and Jean-Baptiste Alayrac and Rohan Anil and Dmitry and Lepikhin and Sebastian Borgeaud and Olivier Bachem and Armand Joulin and Alek Andreev and Cassidy Hardin and Robert Dadashi and Léonard Hussenot},
      year={2025},
      eprint={2503.19786},
      archivePrefix={arXiv},
      primaryClass={cs.CL},
      url={https://arxiv.org/abs/2503.19786}, 
}

@misc{gemini25pro_2025,
  author       = {{Google DeepMind / Google AI}},
  year         = {2025},
  title        = {Gemini 2.5 Pro — Model Card},
  howpublished = {\url{https://cloud.google.com/vertex-ai/generative-ai/docs/models/gemini/2-5-pro}},
  note         = {Multimodal reasoning model with 1M-token context, supporting text, audio, image, video and code input. Accessed 2025-11-29}
}

@misc{vaani_whisper_hi_2025,
  author       = {ARTPARK-IISc},
  year         = {2025},
  title        = {whisper-large-v3-vaani-hindi: Fine-tuned Whisper ASR for Hindi},
  howpublished = {\url{https://huggingface.co/ARTPARK-IISc/whisper-large-v3-vaani-hindi}},
  note         = {Trained on ~718 hours of Hindi speech from multiple datasets including Vaani, Gramvaani, IndicVoices, Fleurs and CommonVoice. Accessed 2025-11-29}
}

@misc{vaani_whisper_kn_2025,
  author       = {ARTPARK-IISc},
  year         = {2025},
  title        = {whisper-medium-vaani-kannada: Fine-tuned Whisper ASR for Kannada},
  howpublished = {\url{https://huggingface.co/ARTPARK-IISc/whisper-medium-vaani-kannada}},
  note         = {Part of the VAANI-Whisper collection for Indic languages (Kannada). Accessed 2025-11-29}
}

@misc{omnilingualasrteam2025omnilingualasropensourcemultilingual,
      title={Omnilingual ASR: Open-Source Multilingual Speech Recognition for 1600+ Languages}, 
      author={Omnilingual ASR team and Gil Keren and Artyom Kozhevnikov and Yen Meng and Christophe Ropers and Matthew Setzler and Skyler Wang and Ife Adebara and Michael Auli and Can Balioglu and Kevin Chan and Chierh Cheng and Joe Chuang and Caley Droof and Mark Duppenthaler and Paul-Ambroise Duquenne and Alexander Erben and Cynthia Gao and Gabriel Mejia Gonzalez and Kehan Lyu and Sagar Miglani and Vineel Pratap and Kaushik Ram Sadagopan and Safiyyah Saleem and Arina Turkatenko and Albert Ventayol-Boada and Zheng-Xin Yong and Yu-An Chung and Jean Maillard and Rashel Moritz and Alexandre Mourachko and Mary Williamson and Shireen Yates},
      year={2025},
      eprint={2511.09690},
      archivePrefix={arXiv},
      primaryClass={cs.CL},
      url={https://arxiv.org/abs/2511.09690}, 
}

@article{chiu2017_medical_asr,
  author       = {Chiu, Chung-Cheng and Tripathi, Anshuman and Chou, Katherine and Co, Chris and Jaitly, Navdeep and Jaunzeikare, Diana and Kannan, Anjuli and Nguyen, Patrick and Sak, Hasim and Sankar, Ananth and Tansuwan, Justin and Wan, Nathan and Wu, Yonghui and Zhang, Xuedong},
  title        = {Speech recognition for medical conversations},
  journal      = {arXiv preprint arXiv:1711.07274},
  year         = {2017},
  doi          = {10.48550/arXiv.1711.07274},
  url          = {https://arxiv.org/abs/1711.07274},
  note         = {Version v2, June 2018. Accessed 2025-11-29}
}

@article{miner2020_asr_psychotherapy,
  author  = {Miner, Adam S. and Haque, Albert and Fries, Jason A. and Fleming, Scott L. and Wilfley, Denise E. and Wilson, G. Terence and Milstein, Arnold and Jurafsky, Dan and Arnow, Bruce A. and Agras, W. Stewart and Fei-Fei, Li and Shah, Nigam H.},
  title   = {Assessing the accuracy of automatic speech recognition for psychotherapy},
  journal = {NPJ Digital Medicine},
  volume  = {3},
  pages   = {82},
  year    = {2020},
  doi     = {10.1038/s41746-020-0285-8},
  url     = {https://doi.org/10.1038/s41746-020-0285-8}
}

@inproceedings{tran2022_digital_scribes,
  author    = {Tran, Brian D. and Mangu, Ramya and Tai-Seale, Ming and Lafata, Jennifer E. and Zheng, Kai},
  title     = {Automatic Speech Recognition Performance for Digital Scribes: A Performance Comparison Between General-Purpose and Specialized Models Tuned for Patient–Clinician Conversations},
  booktitle = {AMIA Annual Symposium Proceedings},
  pages     = {1072--1080},
  year      = {2022},
  note      = {PMCID: PMC10148344; PMID: 37128439}
}

@article{nakatumba2025_asr_bias_review,
  author       = {Nakatumba-Nabende, Joyce and Kagumire, Sulaiman and Kantono, Caroline and Nabende, Peter},
  title        = {A Systematic Literature Review on Bias Evaluation and Mitigation in Automatic Speech Recognition Models for Low-Resource African Languages},
  journal      = {Proceedings of the ACM (via DOI 10.1145/3769089)},
  year         = {2025},
  doi          = {10.1145/3769089},
  url          = {https://dl.acm.org/doi/10.1145/3769089}
}

@inproceedings{cunningham2025_asr_bias_equity,
  author    = {Cunningham, Jay L. and Adjagbodjou, Adinawa and Basoah, Jeffrey and Jawara, Jainaba and Kadoma, Kowe and Lewis, Aaleyah},
  title     = {Toward Responsible ASR for African American English Speakers: A Scoping Review of Bias and Equity in Speech Technology},
  booktitle = {Proceedings of the Eighth AAAI/ACM Conference on AI, Ethics, and Society (AIES 2025)},
  pages     = {665--678},
  year      = {2025},
  url       = {https://ojs.aaai.org/index.php/AIES/article/view/36580/38718},
  note      = {Accessed 2025-11-29}
}

@misc{denoasr,
      title={DENOASR: Debiasing ASRs through Selective Denoising}, 
      author={Anand Kumar Rai and Siddharth D Jaiswal and Shubham Prakash and Bendi Pragnya Sree and Animesh Mukherjee},
      year={2024},
      eprint={2410.16712},
      archivePrefix={arXiv},
      primaryClass={cs.SD},
      url={https://arxiv.org/abs/2410.16712}, 
}


%% file: references.bib
@misc{eka_care_eka_2024,
	title = {The {Eka} {Medical} {ASR} {Evaluation} {Dataset}},
	url = {https://huggingface.co/datasets/ekacare/eka-medical-asr-evaluation-dataset},
	urldate = {2025-11-18},
	publisher = {Hugging Face},
	author = {Eka Care},
	year = {2024},
}

@article{rai_deep_2024,
	title = {A {Deep} {Dive} into the {Disparity} of {Word} {Error} {Rates} across {Thousands} of {NPTEL} {MOOC} {Videos}},
	volume = {18},
	issn = {2334-0770, 2162-3449},
	url = {https://ojs.aaai.org/index.php/ICWSM/article/view/31390},
	doi = {10.1609/icwsm.v18i1.31390},
	language = {en},
	urldate = {2025-11-19},
	journal = {Proceedings of the International AAAI Conference on Web and Social Media},
	author = {Rai, Anand Kumar and Jaiswal, Siddharth D and Mukherjee, Animesh},
	month = may,
	year = {2024},
	pages = {1302--1314},
}

@misc{javed_svarah_2023,
	title = {Svarah: {Evaluating} {English} {ASR} {Systems} on {Indian} {Accents}},
	shorttitle = {Svarah},
	url = {http://arxiv.org/abs/2305.15760},
	doi = {10.48550/arXiv.2305.15760},
	urldate = {2025-11-19},
	publisher = {arXiv},
	author = {Javed, Tahir and Joshi, Sakshi and Nagarajan, Vignesh and Sundaresan, Sai and Nawale, Janki and Raman, Abhigyan and Bhogale, Kaushal and Kumar, Pratyush and Khapra, Mitesh M.},
	month = may,
	year = {2023},
	note = {arXiv:2305.15760 [cs]},
}

@misc{javed_indicsuperb_2022,
	title = {{IndicSUPERB}: {A} {Speech} {Processing} {Universal} {Performance} {Benchmark} for {Indian} languages},
	shorttitle = {{IndicSUPERB}},
	url = {http://arxiv.org/abs/2208.11761},
	doi = {10.48550/arXiv.2208.11761},
	urldate = {2025-11-19},
	publisher = {arXiv},
	author = {Javed, Tahir and Bhogale, Kaushal Santosh and Raman, Abhigyan and Kunchukuttan, Anoop and Kumar, Pratyush and Khapra, Mitesh M.},
	month = dec,
	year = {2022},
	note = {arXiv:2208.11761 [cs]},
}

@misc{javed_indicvoices_2024,
	title = {{IndicVoices}: {Towards} building an {Inclusive} {Multilingual} {Speech} {Dataset} for {Indian} {Languages}},
	shorttitle = {{IndicVoices}},
	url = {http://arxiv.org/abs/2403.01926},
	doi = {10.48550/arXiv.2403.01926},
	language = {en},
	urldate = {2025-11-19},
	publisher = {arXiv},
	author = {Javed, Tahir and Nawale, Janki Atul and George, Eldho Ittan and Joshi, Sakshi and Bhogale, Kaushal Santosh and Mehendale, Deovrat and Sethi, Ishvinder Virender and Ananthanarayanan, Aparna and Faquih, Hafsah and Palit, Pratiti and Ravishankar, Sneha and Sukumaran, Saranya and Panchagnula, Tripura and Murali, Sunjay and Gandhi, Kunal Sharad and R, Ambujavalli and M, Manickam K. and Vaijayanthi, C. Venkata and Karunganni, Krishnan Srinivasa Raghavan and Kumar, Pratyush and Khapra, Mitesh M.},
	month = mar,
	year = {2024},
	note = {arXiv:2403.01926 [cs]},
}

@inproceedings{tatman_gender_2017,
	address = {Valencia, Spain},
	title = {Gender and {Dialect} {Bias} in {YouTube}'s {Automatic} {Captions}},
	url = {https://aclanthology.org/W17-1606/},
	doi = {10.18653/v1/W17-1606},
	urldate = {2025-11-18},
	booktitle = {Proceedings of the {First} {ACL} {Workshop} on {Ethics} in {Natural} {Language} {Processing}},
	publisher = {Association for Computational Linguistics},
	author = {Tatman, Rachael},
	editor = {Hovy, Dirk and Spruit, Shannon and Mitchell, Margaret and Bender, Emily M. and Strube, Michael and Wallach, Hanna},
	month = apr,
	year = {2017},
	pages = {53--59},
}

@article{just_moving_2025,
	title = {Moving beyond word error rate to evaluate automatic speech recognition in clinical samples: {Lessons} from research into schizophrenia-spectrum disorders},
	volume = {352},
	issn = {0165-1781},
	shorttitle = {Moving beyond word error rate to evaluate automatic speech recognition in clinical samples},
	url = {https://www.sciencedirect.com/science/article/pii/S0165178125003385},
	doi = {10.1016/j.psychres.2025.116690},
	urldate = {2025-11-18},
	journal = {Psychiatry Research},
	author = {Just, Sandra Anna and Elvevåg, Brita and Pandey, Shrankhla and Nenchev, Ivan and Bröcker, Anna-Lena and Montag, Christiane and Morgan, Sarah E},
	month = oct,
	year = {2025},
	pages = {116690},
}

@article{russell_what_2024,
	title = {What automatic speech recognition can and cannot do for conversational speech transcription},
	volume = {3},
	issn = {2772-7661},
	url = {https://www.sciencedirect.com/science/article/pii/S2772766124000697},
	doi = {10.1016/j.rmal.2024.100163},
	number = {3},
	urldate = {2025-11-18},
	journal = {Research Methods in Applied Linguistics},
	author = {Russell, Sam O’Connor and Gessinger, Iona and Krason, Anna and Vigliocco, Gabriella and Harte, Naomi},
	month = dec,
	year = {2024},
	pages = {100163},
}

@article{ciampelli_combining_2023,
	title = {Combining automatic speech recognition with semantic natural language processing in schizophrenia},
	volume = {325},
	issn = {0165-1781},
	url = {https://www.scopus.com/pages/publications/85160097289},
	doi = {10.1016/j.psychres.2023.115252},
	number = {115252},
	urldate = {2025-11-18},
	journal = {Psychiatry Research},
	author = {Ciampelli, S. and Voppel, A. E. and de Boer, J. N. and Koops, S. and Sommer, I. E.C.},
	month = jul,
	year = {2023},
}

@article{kuligowska_challenges_2023,
	series = {27th {International} {Conference} on {Knowledge} {Based} and {Intelligent} {Information} and {Engineering} {Sytems} ({KES} 2023)},
	title = {Challenges of {Automatic} {Speech} {Recognition} for medical interviews - research for {Polish} language},
	volume = {225},
	issn = {1877-0509},
	url = {https://www.sciencedirect.com/science/article/pii/S1877050923012590},
	doi = {10.1016/j.procs.2023.10.101},
	urldate = {2025-11-18},
	journal = {Procedia Computer Science},
	author = {Kuligowska, Karolina and Stanusch, Maciej and Koniew, Marek},
	month = jan,
	year = {2023},
	pages = {1134--1141},
}

@article{shikino_clinical_2023,
	title = {Do clinical interview transcripts generated by speech recognition software improve clinical reasoning performance in mock patient encounters? {A} prospective observational study},
	volume = {23},
	issn = {1472-6920},
	shorttitle = {Do clinical interview transcripts generated by speech recognition software improve clinical reasoning performance in mock patient encounters?},
	url = {https://doi.org/10.1186/s12909-023-04246-9},
	doi = {10.1186/s12909-023-04246-9},
	language = {en},
	number = {1},
	urldate = {2025-11-18},
	journal = {BMC Medical Education},
	author = {Shikino, Kiyoshi and Tsukamoto, Tomoko and Noda, Kazutaka and Ohira, Yoshiyuki and Yokokawa, Daiki and Hirose, Yuta and Sato, Eri and Mito, Tsutomu and Ota, Takahiro and Katsuyama, Yota and Uehara, Takanori and Ikusaka, Masatomi},
	month = apr,
	year = {2023},
	pages = {272},
}
